\documentclass[10pt,twocolumn,letterpaper]{article}

\usepackage{wacv}              

\usepackage{graphicx}
\usepackage{amsmath}
\usepackage{amssymb}
\usepackage{booktabs}

\usepackage{multirow}
\usepackage{threeparttable}
\usepackage{bm}
\usepackage[table,xcdraw]{xcolor}

\usepackage[capitalize]{cleveref}
\crefname{section}{Sec.}{Secs.}
\Crefname{section}{Section}{Sections}
\Crefname{table}{Table}{Tables}
\crefname{table}{Tab.}{Tabs.}

\begin{document}

\title{3D WholeBody Pose Estimation based on Semantic Graph Attention Network and Distance Information}

\author{Sihan Wen\textsuperscript{*} \quad Xiantan Zhu \quad Zhiming Tan\\
Fujitsu R\&D Center Co., Ltd.\\
{\tt\small \{wensihan, zhuxiantan, zhmtan\}@fujitsu.com}
}
\maketitle

\begin{abstract}
In recent years, a plethora of diverse methods have been proposed for 3D pose estimation. Among these, self-attention mechanisms and graph convolutions have both been proven to be effective and practical methods. Recognizing the strengths of those two techniques, we have developed a novel Semantic Graph Attention Network which can benefit from the ability of self-attention to capture global context, while also utilizing the graph convolutions to handle the local connectivity and structural constraints of the skeleton. We also design a Body Part Decoder that assists in extracting and refining the information related to specific segments of the body. Furthermore, our approach incorporates Distance Information, enhancing our model's capability to comprehend and accurately predict spatial relationships. Finally, we introduce a Geometry Loss who makes a critical constraint on the structural skeleton of the body, ensuring that the model's predictions adhere to the natural limits of human posture. The experimental results validate the effectiveness of our approach, demonstrating that every element within the system is essential for improving pose estimation outcomes. With comparison to state-of-the-art, the proposed work not only meets but exceeds the existing benchmarks.
\end{abstract}
\begin{figure*}[ht]
	\footnotesize
	\begin{center}
		\includegraphics[width=0.8\linewidth]{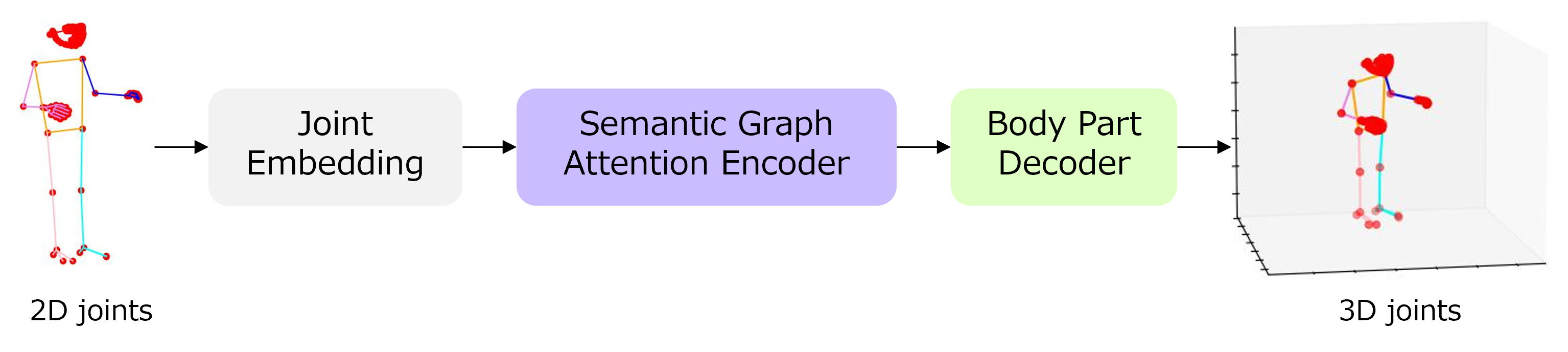}
		\caption{Overview of proposed method.}
		\label{fig:1}
	\end{center}
\end{figure*}
\section{Introduction}
3D human pose estimation is a critical task in computer vision that involves identifying and locating key points of the human body within images or video frames. This capability is essential for a wide range of applications, including analyzing human behavior, understanding intentions, and studying how individuals interact and communicate with their physical environment. Accurate pose estimation can provide insights into activities of daily living, sports analysis, virtual and augmented reality, and human-computer interaction.

Currently, two primary directions are emerged for 3D pose estimation. The first approach directly estimates the 3D body pose from a single RGB image, utilizing advanced deep learning techniques and models such as \cite{SingleShotM3, LCRNetM2, Monocular3H}. The second approach employs a two-stage process, where 2D keypoint estimation is followed by a lift from the 2D human pose to a 3D space, as demonstrated in works like \cite{SimpleBaseline, CanonPoseSM}. 

Since 2D pose estimators are already optimized for speed and accuracy \cite{cao2017realtime, sun2019deep}, focusing on 2D to 3D lifting process based on 2D pose estimation achievements can significantly reduce development time and computational overhead. SimpleBaseline\cite{SimpleBaseline} provide an initial straightforward approach for pose lifting employed a multi-layer perceptron, which remarkably achieved high precision despite the absence of image feature data. However, subsequent research has pointed out that considering pose as a mere vector overlooks the spatial interconnections between the body's joints. Then, the Graph Convolutional Networks (GCNs)\cite{SpatialTG, GraphSH} was proposed to directly deal with a general class of graphs. Among those GCNs methods, semantic graph convolutions networks (SemGCN) does not rely on hand-crafted constraints to analyze the patterns for a specific application, and significantly improves the power of graph convolutions. On the other side, transformer-based \cite{JointformerSL, EndtoEndHP} approaches that use the more generalized self-attention mechanism to learn these relationships within a sequence of tokens representing joints shows a considerable success.

Therefore, we propose a novel network called Semantic Graph Attention Network (SemGAN), which combines SemGCN and Self-attention together to obtain the global features from Self-attention and local features based on the prior information from graph. 

Another critical issue for many lifting methods in the field of 3D human pose estimation is that they rely on datasets that focused on specific parts of the body: datasets for the body joints such as \cite{h36m, COCO, PanopticSA}, facial landmarks like \cite{Pix2FaceD3, LargeP3}, and hand poses including \cite{3DHS, Modelbased3H}, leading to complex training pipelines and heterogeneous evaluations. The introduction of Human3.6M 3D WholeBody (H3WB) dataset has significantly addressed the limitations of previous datasets by providing comprehensive 3D annotations of the whole-body including body, face, hands. Our work is conducted on the H3WB dataset, aiming to harness its rich and detailed 3D annotations to achieve outstanding performance in the field of 3D human pose estimation.

Our contributions can be summarized as follows:
\begin{itemize}
	\item We embed the SemGCN into Self-Attention method to construct a SemGAN, which helps to obtain the local and non-local features among 133 joints.
	\item We segment the whole-body joints into three individual parts: body, face, and hands, which allows us to effectively leverage the neighboring joints with high correlation coefficients.
	\item We add the distance information between each joint and its parent joint to better understand the relative positions of body parts, and apply a geometry loss for training process, including the normal loss and bone loss.
	\item We achieve the first place on H3WB benchmark, exceeding the second place by 15.44 in Mean Per Joint Position Error (MPJPE) measurement.
\end{itemize}
\section{Related Works}
\textbf{Attention Transformer}
Attention Mechanisms initially emerged in computer vision, with pivotal work in \cite{mnih2014recurrent} demonstrating how models could focus on relevant parts of the input data. The adoption of attention in NLP by Bahdanau et al.\cite{bahdanau2014neural} marked a turning point, offering a new paradigm for sequence-to-sequence tasks. The Transformer model, introduced in the seminal paper by Vaswani et al.\cite{AttentionIA}, has become a cornerstone in modern deep learning. Its encoder-decoder structure, multi-head attention, and positional encoding have set a new standard for handling sequential data without the constraints of traditional RNNs and CNNs. The impact and development of Transformer models have been profound, with models like BERT and GPT showcasing their versatility and effectiveness in a myriad of NLP tasks. In computer vision, the adaptation of Transformers has given rise to architectures such as the Vision Transformer, which has demonstrated competitive performance in image classification and pose estimation. For example, \cite{JointformerSL, EndtoEndHP, Zheng20213DHP} all achieve an impressive results on 3D pose estimation based on the attention mechanisms.

\textbf{Semantic Graph Convolutions}
Graph Convolutional Networks (GCNs) have emerged as a powerful tool for processing graph-structured data, with applications ranging from social network analysis to molecule modeling. However, traditional GCNs\cite{SpatialTG, GraphSH, Yang2018GraphRF, Kipf2016SemiSupervisedCW} suffer from limited expressiveness due to their reliance on fixed neighborhood aggregations and shared transformation matrices across all nodes. To address these limitations, researchers have proposed SemGCN\cite{SemGCN}, which learn to capture implicit semantic information such as local and global node relations, enhancing the capability of GCNs. Additionally, self-supervised learning has been integrated into GCNs to align node features from different perspectives, further improving performance on classification or regression.

\section{Methods}
In this section, we describe the proposed architecture for 3D whole-body estimation, as shown in \cref{fig:1}. Based on the procedure of JointFormer \cite{JointformerSL}, we first feed the 2D joints into Joint Embedding layer to get a higher dimension features, then design a novel Semantic Graph Attention Encoder and Body Part Decoder model to better estimate the 3D whole-body pose.
\subsection{Semantic Graph Attention Encoder}
Self-attention mechanisms can capture a comprehensive representation of the global content by computing attention scores across the entire features. And, SemGCN by learning input-independent weights for edges which represent priors implied in the graph structures, can extract a richer set of local information. Considering the notable strengths of these two models, we craft a Semantic Graph Attention Encoder model through the combination of this two networks.
\begin{figure}[ht!]
	\footnotesize
	\begin{center}
		\includegraphics[width=1.0\linewidth]{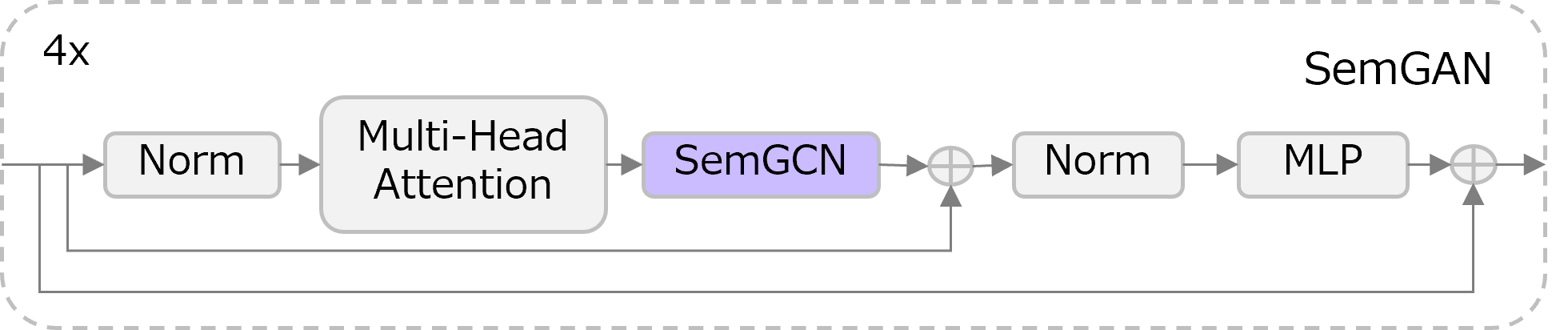}
		\caption{Semantic Graph Attention Encoder.}
		\label{fig:2}
	\end{center}
\end{figure}

The pipeline of Semantic Graph Attention Encoder is designed as \cref{fig:2}, consists of four layers of SemGAN. We embed the SemGCN in the last layer of self-attention which helps to maintain the global information from self-attention and equip with local information from SemGCN.

\subsection{Body Part Decoder}
\label{sec:BPD}
Due to the mass and density of keypoints present in the human face and hands, we separate the features into there individual parts in decoder, which can better leverage the neighboring joints with high correlation coefficients within each part. 

\begin{figure}[ht]
	\footnotesize
	\begin{center}
		\includegraphics[width=1.0\linewidth]{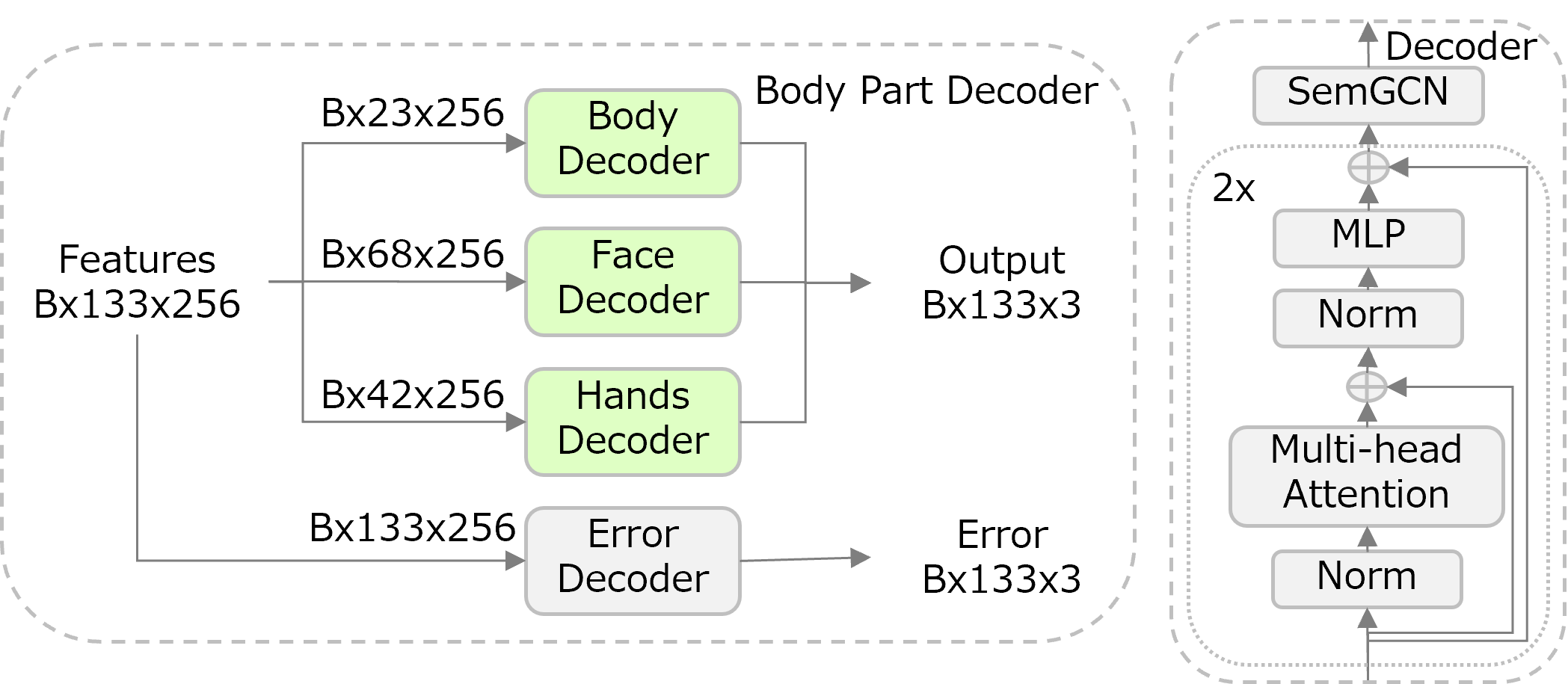}
		\caption{Body Part Decoder. B stands for batch size, 133 is the number joints of whole-body, 256 is the dims of feature, 23 is number joints of body, 68 is number joints of face, 42 is number joints of hands.}
		\label{fig:3}
	\end{center}
\end{figure}

The framework of Body Part Decoder is designed as \cref{fig:3}. We use the self-attention network by twice in decoder to capture more correlated features within different body part. And endorse the SemGCN at the last layer instead of traditional linear which helps to get accurate location with assistance information from adjacency.
\subsection{Distance Information}
To explore the spatial relationships inherent in skeletal structures, we provide a data process method to obtain the distance between every joint and its parent. With this additional information a deeper sense of spatial orientation and relative positioning can be developed.
 
The inputs and distance is defined as:
\begin{gather}
X_{i} = \left(J_{i}, D_{i}\right)\\
D_{i} = \|\bm{J}_{i} - \bm{J}_{\tilde{i}}\|
\end{gather}
$i=0,1,...,132$, ${\tilde{i}}$ represents the associated parent of ${i}$, ${D_{i}}$ is the Euclidean Distance calculated by current joints ${J_{i}}$ and is parent joints ${J}_{\tilde{i}}$. The inputs is finally concatenated by the 2D inputs and distance to get $X_{(x, y, d)}$.

\subsection{Loss Function}
According to \cite{JointformerSL} demonstrated, with the addition of the error prediction, the training can be stabilized and leads to better overall result, we also utilize the error prediction in our loss. Furthermore, we add another two effectiveness loss during our training, normal loss and bone loss.
\begin{equation}
	\mathcal{L}_{all}=\sum{\alpha\mathcal{L}_{3D} + \beta\mathcal{L}_{error} + \gamma\mathcal{L}_{normal} + \delta\mathcal{L}_{bone}}
\end{equation}	
${\mathcal{L}_{3D}}$ is the mean-squared error between 3D prediction and ground-truth. ${\mathcal{L}_{error}}$ is the mean-squared error between prediction error and true error. The true error is defined as the absolute difference between the prediction and the ground-truth ${\hat{e}=|y-\hat{y}|}$.
${\alpha, \beta, \gamma, \delta}$ are the weights of different loss. Through the observation of training loss and attempt of training trails, we set them to ${5e-1, 2.5e-1, 2.5e-4, 5e-4}$, respectively.

\textbf{Normal Loss.}
Normal loss is first raised by \cite{surface_normal}, and has been confirmed to be practical in \cite{virtual_normal, matsune2024geometry}. We following the strategy in \cite{matsune2024geometry}, choosing the most powerful four part of body: ${L_{arm}}$, ${R_{arm}}$ , ${L_{leg}}$, ${R_{leg}}$. With those triangles normal vector, the loss function can be defined as:
\begin{equation}
\mathcal{L}_{normal}=\frac{1}{M}\left(\sum_{i=1}^{M}\left\|\bm{n}_{i}^{g t}-\bm{n}_{i}^{\text {pred}}\right\|_{1}\right)
\end{equation}
where $M$ denotes the number of triangles, here we set it to 4 corresponding to the selected body parts. ${n}$ stands for the normal vector of the local plane.

\textbf{Bone Loss.}
Bone loss as a constraints in human poses estimation is also widely be used \cite{Sun_2017, Zhou_2017, Pavlakos_2017, Habibie_2019, matsune2024geometry}. We adopt this loss to deliver a limitation on skeleton and echo with the additional input distance information. The bone loss is defined as:

\begin{equation}
	\mathcal{L}_{bone}=\frac{1}{L}\left(\sum_{i=1}^{L}\left\|\bm{b}_{i}^{g t}-\bm{b}_{i}^{\text {pred}}\right\|_{1}\right)
\end{equation}
where ${L}$ is the number of bone vector, we use all the 133 joints. Each bone ${b}$ is a directed vector pointing from the starting joint to its associated parent.
\begin{table*}[htbp]
	\centering
	\footnotesize
	\renewcommand\arraystretch{1.3}
	\begin{tabular}{lcccc}\hline
		Method & All & Body & Face/Aligned & Hands/Aligned \\\hline\hline
		JointFormer\textsuperscript{*} & 67.50 & 59.24 & 48.64/10.10 & 101.45/29.41 \\
		Semantic Graph Attention Encoder & 53.90 & 48.27 & 44.49/17.66 & 72.23/27.16 \\
		Semantic Graph Attention Encoder-Body Part Decoder & 52.57 & 46.82 & 43.75/17.08 & 69.66/26.78 \\
		Semantic Graph Attention Encoder-Body Part Decoder-Distance & 49.97 & 46.24 & 41.23/24.48 & \textbf{66.16}/\textbf{25.87} \\
		Semantic Graph Attention Encoder-Body Part Decoder-Distance-Loss & \textbf{47.87} & \textbf{45.39} & \textbf{36.37}/\textbf{15.95} & 67.86/27.77\\\hline
	\end{tabular}
	\caption{\label{exp}Ablation study of the effect on different part. JointFormer\textsuperscript{*} represents for the network reproduced by ourselves.}
\end{table*}
\section{Experiments}
In this section, we first introduce the datasets to train and evaluate our proposed approach, then show the implementation details, and finally perform an ablation study and a comparisons with state-of-the-art methods.

\subsection{Datasets and Evaluation}
Our proposed approach is evaluated on H3WB \cite{H3WB} dataset, a recent benchmark for diverse whole-body pose estimation tasks.

\subparagraph{Datasets.}
H3WB is a new large-scale dataset for accurate 3D whole-body pose estimation, which extend from Human3.6M \cite{h36m} with 3D keypoint annotations. It consists of 133 paired 2D and 3D whole-body keypoint annotations for a set of 100k images from Human3.6M, following the same layout used in COCO WholeBody \cite{jin2020wholebody}.
The training set we used are samples from S1, S5, S6 and S7, including 80k \{image, 2D, 3D\} triplets. The test set contains all samples from S8, including 20k triplets.

\subparagraph{Evaluation.}
For evaluation, we following the metric defined in \cite{H3WB}:
\begin{itemize}
	\item MPJPE for the whole-body, the body, the face and the hands when all joints are aligned with the pelvis, which in our case is the middle of left and right hip,
	\item MPJPE for the face when it is centered on the nose,
	\item MPJPE for the hands when hands are centered on the wrists, i.e left hand aligned with keypoint 92 and right hand aligned with keypoint 113.
\end{itemize}

\subsection{Implementation Details}
We implement our method in PyTorch based on NVIDIA A40. The network is trained for 100 epochs with a batch size of 200 using AdamW~\cite{adamw} optimizer. We set the initial learning rate to 0.001 and using a cosine annealing learning rate decay \cite{Loshchilov2016SGDRSG}. And we use test-time data augmentation by horizontal flipping, following \cite{Pavllo20183DHP}.

\subsection{Ablation Studies}
To ascertain the impact of each part on the model's predictive power, we conducted an ablation study. \cref{exp} shows the overview of our ablation studies.

Examining the comprehensive results from \cref{exp}, we find that every element plays a role in enhancing the precision of 3D whole-body pose estimation. The most substantial improvement is observed in Semantic Graph Attention Encoder part, which significantly boosting precision by integrating more global and local features. Following that is the additional Distance Information in the input, which provides a effective improvement for areas of the body that have a higher concentration of features, such as the face and hands. Subsequently, under the constraints of Normal and Bone loss, the model's accuracy is further enhanced. Although the overall improvement from the Body Part Decoder section is not substantial, the accuracy of body part is notably improved due to the separation of the whole-body into parts with varying densities.

\subsection{Comparison to the State-of-the-Art Methods}
We compare the performance against the three state-of-the-art 3D whole-body pose estimation methods, such as Large SimpleBaseline~\cite{SimpleBaseline}, JointFormer~\cite{JointformerSL}, and 3D-LFM~\cite{3DLFMLF}. The results are presented in \cref{sota}.
\begin{table}[h!]
	\centering
	\footnotesize
	\scalebox{0.9}{
	\renewcommand\arraystretch{1.3}
	    \begin{tabular}{lcccc}\hline
	    	Method & All & Body & Face/Aligned & Hand/Aligned \\\hline\hline
	    	Large SimpleBaseline & 112.3 & 112.6 & 110.6/14.6 & 114.8/31.7 \\
	    	JointFormer & 88.3 & 84.9 & 66.5/17.8 & 125.3/43.7 \\
	    	3D-LFM & 64.13 & 60.83 & 56.55/\textbf{10.44} & 78.21/28.22 \\
	    	Proposed & \textbf{47.87} & \textbf{45.39} & \textbf{36.37}/15.95 & \textbf{67.86}/\textbf{27.77} \\\hline
	    \end{tabular}
    } 
	\caption{\label{sota}Comparison to the state-of-the-art methods on H3WB test dataset. Results are calculated in millimeters for MPJPE metric.}
\end{table}

 As demonstrated by the experimental comparison, our approach significantly outperforms the current state-of-the-art techniques, achieving a substantial reduction in the MPJPE for the body part, decreasing from 60.83mm to 45.39mm.

\section{Conclusion}
Our proposed system incorporates several innovative components that collectively enhance the performance of human pose estimation and understanding: A Semantic Graph Attention Encoder that effectively harnesses both global and local features; A Body Part Decoder which introduced to obtain more detailed insights specific to each corresponding body part; A Distance Information who can enhance the model's ability to understand the relative positions and spatial relationships between related joints; and A Geometry Loss is implied to constrain and maintain the structural integrity of the body skeleton. Overall, these advancements delivering highly detailed and accurate 3D whole-body pose estimations.

\clearpage
{\small
\bibliographystyle{ieee_fullname}
\bibliography{egbib}
}

\end{document}